\documentclass[sigconf]{acmart}
\AtBeginDocument{%
  }
\usepackage{booktabs}
\usepackage{makecell}
\usepackage{pifont}
\usepackage{xcolor}
\usepackage{graphicx}
\usepackage{placeins}
\usepackage[capitalize]{cleveref}
\usepackage{array}

\definecolor{markgreen}{HTML}{2E7D32}
\definecolor{markorange}{HTML}{D97706}
\definecolor{markred}{HTML}{C62828}
\newcommand{\cmark}{\textcolor{markgreen}{\ding{51}}}
\newcommand{\pmark}{\textcolor{markorange}{\ensuremath{\triangle}}}
\newcommand{\xmark}{\textcolor{markred}{\ding{55}}}
\newcommand{\finlifebench}{\textsc{FinLifeBench}}

\setcopyright{none}
\copyrightyear{2026}
\acmYear{2026}
\acmDOI{}
\acmConference[ICAIF '26]{7th ACM International Conference on AI in Finance}%
  {November 14--17, 2026}{Milan, Italy}
\acmISBN{}

\begin{document}

\title{FinLifeBench: Exhaustive Life-Event History and Financial-State Reconstruction from Longitudinal Banking Dialogue}

%%
%% Authors are listed in contribution order. The first two authors
%% contributed equally; Pilsung Kang is the corresponding author.
\author{Hangyeul Lee}
\authornote{Both authors contributed equally to this research.}
\email{mikelee@snu.ac.kr}
\orcid{0009-0000-6948-1533}
\affiliation{%
  \institution{Seoul National University}
  \city{Seoul}
  \country{Republic of Korea}
}

\author{Juyoung Oh}
\authornotemark[1]
\email{aven.j@lab.kakaobank.com}
\orcid{0009-0000-9858-5315}
\affiliation{%
  \institution{KakaoBank}
  \department{Financial Tech Lab}
  \city{Seongnam}
  \country{Republic of Korea}
}

\author{Jaeyong Ko}
\email{jyko22@snu.ac.kr}
\orcid{0009-0006-8319-8316}
\affiliation{%
  \institution{Seoul National University}
  \city{Seoul}
  \country{Republic of Korea}
}

\author{Sunmin Kim}
\email{sunmin\_kim@snu.ac.kr}
\orcid{0009-0006-8995-6157}
\affiliation{%
  \institution{Seoul National University}
  \city{Seoul}
  \country{Republic of Korea}
}

\author{Jaeik Park}
\email{jaeiksan@snu.ac.kr}
\orcid{0009-0006-2881-0197}
\affiliation{%
  \institution{Seoul National University}
  \city{Seoul}
  \country{Republic of Korea}
}

\author{Hyunkyu Kim}
\email{conor.k@lab.kakaobank.com}
\orcid{0009-0005-7858-8136}
\affiliation{%
  \institution{KakaoBank}
  \department{Financial Tech Lab}
  \city{Seongnam}
  \country{Republic of Korea}
}

\author{Jungmin Son}
\email{elena.son@lab.kakaobank.com}
\orcid{0009-0003-3999-1892}
\affiliation{%
  \institution{KakaoBank}
  \department{Financial Tech Lab}
  \city{Seongnam}
  \country{Republic of Korea}
}

\author{Pilsung Kang}
\authornote{Corresponding author.}
\email{pilsung\_kang@snu.ac.kr}
\orcid{0000-0001-7663-3937}
\affiliation{%
  \institution{Seoul National University}
  \city{Seoul}
  \country{Republic of Korea}
}

\renewcommand{\shortauthors}{Lee, Oh, et al.}

\begin{abstract}
Repeated banking interactions require assistants to maintain complete, current, and traceable customer records as life changes emerge incidentally in routine requests. Existing benchmarks emphasize question answering, bounded episodes, or targeted recall rather than exhaustive longitudinal reconstruction. \finlifebench\ evaluates two tasks over the same cumulative dialogue: reconstructing every life-event instance with its first-establishing session and reconstructing a complete 34-path financial state at consecutive checkpoints. The benchmark contains 6,000 eight-turn Korean banking sessions from 20 independent synthetic trajectories, with deterministic, exhaustive gold for 24 event types and 34 state paths. Quality assurance combines deterministic validation with semantic review. Across eleven LLMs under a full-context condition without persistent memory across requests, event–anchor recall falls from 0.591 at 15 sessions to 0.445 at 300. Errors are driven primarily by omitted events rather than poor anchor localization, while financial-state reconstruction frequently treats superseded or potentially outdated information as current; the best GCA@15 reaches 0.470. These results show that models can localize evidence for recovered events while still failing to maintain complete and temporally valid longitudinal records.
\end{abstract}

\begin{CCSXML}
<ccs2012>
   <concept>
       <concept_id>10010147.10010178.10010179.10010181</concept_id>
       <concept_desc>Computing methodologies~Discourse, dialogue and pragmatics</concept_desc>
       <concept_significance>500</concept_significance>
       </concept>
   <concept>
       <concept_id>10010405.10003550.10003556</concept_id>
       <concept_desc>Applied computing~Online banking</concept_desc>
       <concept_significance>300</concept_significance>
       </concept>
 </ccs2012>
\end{CCSXML}

\ccsdesc[500]{Computing methodologies~Discourse, dialogue and pragmatics}
\ccsdesc[300]{Applied computing~Online banking}

\keywords{longitudinal dialogue, conversational memory, life-event history reconstruction,
financial-state reconstruction, benchmark}

\maketitle

\section{Introduction}

Conversational interfaces are becoming increasingly common in digital finance \citep{BankingDoneRight,ComprehensiveFF}. Across repeated interactions, routine requests may incidentally reveal life events that alter financial needs and invalidate recorded information \citep{GettingToKnowYou}. A longitudinal assistant must identify these changes, retain their evidence, and update the customer's state without relying on stale assumptions \citep{TowardLifelongDialogueAgents,ES-MemEval,KeepMeUpdated}. In finance, failures have direct service consequences: missed changes may suppress relevant support or recommendations, while stale household, employment, housing, or financial-obligation records can yield unsuitable guidance and contradictory treatment across later interactions. Reliable systems must therefore recover both what changed and which financial facts remain up-to-date, with evidence that makes each update traceable. Single-session success cannot establish this complete, current, and traceable record maintenance \citep{Memora,BeyondGoldfishMemory}.

Yet existing benchmarks largely leave this requirement untested. They emphasize financial question answering and retrieval \citep{FinQA,ConvFinQA,TAT-QA}, request-level understanding \citep{banking77,FIAD}, bounded state tracking \citep{MultiWOZ}, or targeted memory retrieval \citep{LoCoMo,LongMemEval,MemoryAgentBench,DynamicMem}. Prior work establishes state-first generation, provenance-aware evolution, direct hidden-state recovery, and lifecycle- or conflict-level memory diagnosis. To our knowledge, no prior benchmark requires \emph{generative, schema-complete reconstruction} of both a grounded life-event history and the corresponding structured financial state at repeated checkpoints.

% Conversational interfaces are becoming increasingly common in digital finance \citep{ChatbotsinBanking,BankingDoneRight,ComprehensiveFF}. Across repeated interactions, routine requests may incidentally reveal life events that alter financial needs and invalidate recorded information \citep{GettingToKnowYou}. A longitudinal assistant must identify these changes, retain their evidence, and update the customer's state without relying on stale assumptions \citep{TowardLifelongDialogueAgents,ES-MemEval,KeepMeUpdated}. Single-session success therefore cannot establish complete, current, and traceable record maintenance \citep{Memora,BeyondGoldfishMemory}.

% Existing benchmarks instead emphasize financial question answering and retrieval \citep{FinQA,ConvFinQA,TAT-QA,PACIFIC,KoBankIR,FinDER}, request-level understanding \citep{banking77,FIAD}, bounded state tracking \citep{MultiWOZ,TowardsSM}, or targeted memory retrieval \citep{LoCoMo,LongMemEval,MemoryAgentBench,DynamicMem,EvolMem}. None requires checkpoint-wise reconstruction of both a complete, grounded life-event history and the corresponding structured financial state.

\begin{figure*}[t]
\centering
\includegraphics[width=\textwidth]{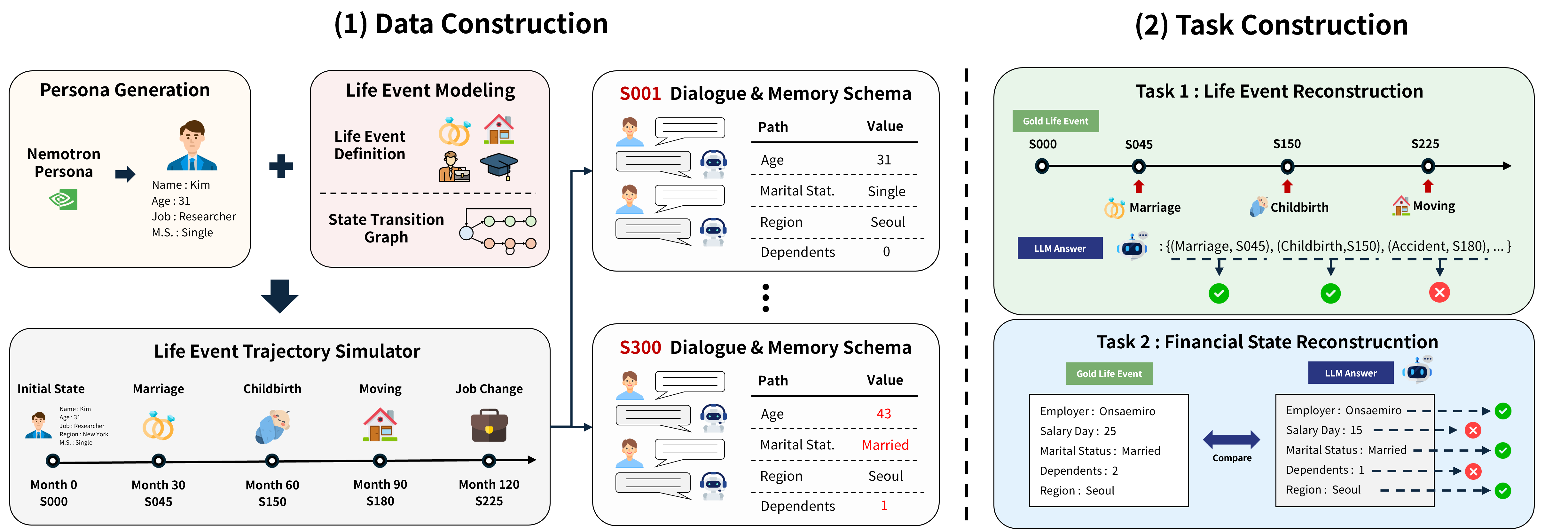}
\caption{Overview of \finlifebench\ construction and evaluation. A persona-conditioned life-event trajectory determines the longitudinal dialogues and financial-state histories. Task~1 reconstructs all event--anchor pairs, whereas Task~2 reconstructs the complete financial state at each checkpoint.} \Description{The figure has two parts. Data construction combines a Korean persona with life-event definitions and a state-transition graph, simulates a life-event trajectory, and generates longitudinal dialogues with corresponding financial-state records. Task construction evaluates recovery of event--anchor pairs in Task 1 and complete financial-state values in Task 2.}
\label{fig:benchmark-overview}
\end{figure*}

To address this gap, we introduce \finlifebench, a benchmark for information reconstruction in longitudinal financial dialogue. Each benchmark trajectory follows a curated, persona-conditioned sequence of life events, yielding deterministic gold annotations: a known life-event history, financial states, and corresponding evidence provenances. We define two reconstruction tasks from these annotations. Task~1 asks a model to reconstruct life-event history along with each first-establishing session. Task~2 asks it to reconstruct the full financial state across 34 state paths. At each checkpoint, a fresh request receives the complete dialogue prefix; persistent-memory storage, updating, and retrieval policies are outside the evaluation scope. \Cref{fig:benchmark-overview} gives an overview of the framework.

Across eleven LLMs given the complete dialogue history available at each checkpoint, we find that the main challenge is maintaining a complete and temporally valid record. For life-event history reconstruction, models increasingly omit previously established events as the dialogue grows, even though they usually identify the correct anchor session for life events they had reconstructed. For financial-state reconstruction, models frequently treat superseded or potentially outdated information as still current. The two tasks also yield different model rankings, highlighting the value of evaluating both reconstruction objectives.

Our contributions are threefold:
\begin{itemize}
\item We introduce \finlifebench, a benchmark of 6,000 Korean banking sessions for reconstructing cumulative life-event history and financial state from long multi-session banking dialogues, with metrics for completeness, provenance, and temporal validity.

\item We formulate two schema-complete reconstruction tasks over the same cumulative dialogue: (1) reconstructing every occurred life-event instance with its first-establishing session and (2) reconstructing all 34 financial-state paths at consecutive checkpoints.

\item Through a systematic evaluation of eleven LLMs, we show that longitudinal reconstruction failures are not captured by a single notion of accuracy: incompleteness, evidence localization, and temporal-state maintenance behave differently across the two reconstruction objectives.
\end{itemize}

% Our contributions are fourfold:
% \begin{itemize}
% \item We introduce \finlifebench, a benchmark for reconstructing cumulative life-event history and financial state from long multi-session banking dialogues, with metrics for completeness, provenance, and temporal validity.
% \item We formulate two exhaustive reconstruction tasks over the same cumulative dialogue: (1) reconstructing every life-event instance with its first-establishing session and (2) reconstructing the complete 34-path financial state at successive checkpoints.
% \item We suggest a metric suite that separates completeness, provenance, and temporal validity, and show these views are not interchangeable: aggregate snapshot accuracy is dominated by unchanged cells, so a dialogue-blind \texttt{Initial Copy} baseline reaches 0.669 CSA yet only 0.177 GCA@15.
% \item Across eleven LLMs, failure patterns recur across model families but differ across objectives: on event--anchor pairs every model leaves more gold pairs unmatched than it emits unmatched predictions (missed:spurious $\geq$ 1.6:1), whereas aggregate state-transition error counts are dominated by spurious updates. \texttt{historical} and \texttt{stale} statuses are frequently predicted as \texttt{current}, conditional anchor accuracy reaches 86.6\% once the event type is recovered, and performance is only weakly associated across the two objectives.
% \end{itemize}

\section{Related Work}

Long-horizon benchmarks cover recall, temporal reasoning, incremental updating, and evolving profiles. \Cref{tab:benchmark-comparison} compares representative works. Beyond these entries, MINTEval studies interference and forgetting, while CloneMem uses top-down coherent trajectories for evolving personal states \citep{MINTEval,CloneMem}.

\begin{table}[t]
  \centering
  \scriptsize
  \setlength{\tabcolsep}{0.8pt}
  \renewcommand{\arraystretch}{1.04}
  \caption{Coverage of benchmark design and evaluation properties.
  \cmark\ denotes direct coverage, \pmark\ related but indirect coverage,
  and \xmark\ no explicit coverage. \emph{Full-state reconstruction}
  requires every schema field to be emitted at each checkpoint;
  benchmarks that query a sampled subset of fields, or that evaluate by
  multiple choice, receive \pmark\ or \xmark. \emph{Life-event tracking}
  requires events to be an evaluation target, not only a generative
  driver of state change.}
  \label{tab:benchmark-comparison}
  \begin{tabular}{@{}
    >{\raggedright\arraybackslash}m{0.29\columnwidth}
    *{5}{>{\centering\arraybackslash}m{0.125\columnwidth}}@{}}
    \toprule
    & \multicolumn{2}{c}{Data design}
      & \multicolumn{2}{c}{Evaluation target} & Domain \\
    \cmidrule(lr){2-3}\cmidrule(lr){4-5}\cmidrule(l){6-6}
    Benchmark
      & Implicit state-change inference
      & State-first trajectory generation
      & Life-event tracking
      & Full-state reconstruction
      & Financial task \\
    \midrule
    LoCoMo~\citep{LoCoMo} & \pmark & \pmark & \cmark & \xmark & \xmark \\
    LoCoMo-Plus~\citep{LoCoMo_Plus} & \cmark & \xmark & \xmark & \xmark & \xmark \\
    MemoryAgentBench~\citep{MemoryAgentBench} & \xmark & \xmark & \xmark & \xmark & \xmark \\
    LongMemEval~\citep{LongMemEval} & \pmark & \xmark & \pmark & \xmark & \xmark \\
    PersonaMem~\citep{PersonaMem} & \cmark & \pmark & \pmark & \xmark & \xmark \\
    AMemGym~\citep{AMemGym} & \cmark & \cmark & \xmark & \pmark & \xmark \\
    HorizonBench~\citep{HorizonBench} & \cmark & \cmark & \pmark & \xmark & \xmark \\
    DynamicMem~\citep{DynamicMem} & \cmark & \cmark & \pmark & \pmark & \xmark \\
    MEMPROBE~\citep{MEMPROBE} & \xmark & \cmark & \xmark & \pmark & \xmark \\
    \midrule
    \textbf{\finlifebench} & \cmark & \cmark & \cmark & \cmark & \cmark \\
    \bottomrule
  \end{tabular}
\end{table}

State-first construction with change-level provenance is by now standard \citep{HorizonBench,DynamicMem,AMemGym}; what still varies is how much of the resulting state an evaluation actually inspects. HorizonBench is closest to our design. It generates dialogues from a structured mental-state graph and records
the triggering event behind every preference change, and it reports that models across families anchor on pre-evolution values \citep{HorizonBench}. DynamicMem also builds state first, but keeps the traces implicit and scores profile completion at checkpoints \citep{DynamicMem}. AMemGym predefines both profiles
and state-evolution trajectories, then lets the agent interact on-policy \citep{AMemGym}. In each case the model is asked about part of the state: a sampled subset of fields, or a recognition choice among candidates. MemOps and MemConflict narrow the target further still, probing individual lifecycle operations \citep{MemOps} and query-conditioned temporal validity through retrieval-and-ranking diagnostics \citep{MemConflict}. \finlifebench\ asks for the whole record. At every checkpoint of every banking trajectory the model must emit each schema field generatively, and we score all of it at the pair and cell level, with no query to indicate where to look. 

Task~2 has a direct analogue in MEMPROBE, which recovers a hidden
31-dimensional user state from the memory artifact an agent produces after interacting \citep{MEMPROBE}. The input is what separates us: we reconstruct from the dialogue itself, at 20 consecutive checkpoints, and pair the state with a life-event history in which every instance is linked back to the session that
first establishes it.

Retrieval and persistent-memory systems select, consolidate, and overwrite evidence through typically query-conditioned interfaces \citep{RAG,MemGPT,Mem0}. \finlifebench\ instead provides every model with the same static dialogue prefix and evaluates exhaustive reconstruction; adapting these systems would require explicit coverage and state-synchronization mechanisms. Off-policy static prefixes are known to introduce reuse bias relative to on-policy interaction \citep{AMemGym}; we use them so that all eleven models receive identical evidence, while leaving memory-management policies outside the evaluation scope.

\section{\finlifebench\ Dataset}
\label{sec:construction}

\finlifebench\ comprises 20 persona-conditioned trajectories, each with 300 banking sessions, 20 life-event instances, and state defined over a common 34-path schema, for 6,000 sessions in total. \finlifebench\ follows a synthetic pipeline in which personas, initial financial states, and life-event trajectories are constructed before their Korean banking dialogues. This ordering ensures deterministic life-event histories, checkpoint states, and evidence provenance.

\subsection{Persona and Life-Event Trajectory Generation}

We draw 20 seed personas from NVIDIA's Korean Nemotron Personas dataset \citep{NemotronPersonasKorea}. The age-band quotas follow aggregate user proportions from an anonymized service operated by the industry partner: four personas in their 20s, six in their 30s, six in their 40s, and four in their 50s. Raw personas are deterministically normalized into typed demographic, household, employment, housing, financial, and dialogue-style fields, from which we instantiate the initial 34-path state and standing financial actions. State cells distinguish unknown from inapplicable values and retain revision histories.

The life-event ontology and financial-state schema are summarized in \Cref{tab:benchmark-ontologies}. Both focus on persistent, dialogue-recoverable information relevant to financial servicing: Task~1 captures life changes that may alter customer needs or invalidate prior records, whereas Task~2 represents profile fields whose values or validity may change accordingly. Their relationship is many-to-many: one event may update several state paths, and a path may be affected by different events. The two sides of the table are therefore independent inventories rather than row-wise mappings.

We generate life-event trajectories from a finite-state transition graph, which keeps them internally consistent and reproducible across runs. Its edges encode two kinds of constraint: precedence, so that divorce cannot precede marriage, and minimum intervals between dependent events, such as the gap from pregnancy to childbirth. Recurrent events appear as loops, each carrying its own repeat limit and cooldown period. For each persona we select an entry point partway through the graph, so that trajectories begin mid-life rather than at a canonical origin, sample a subgraph reachable from that point, and linearize it into an ordered event sequence. The simulator then re-checks every transition against the evolving state and applies each admitted event to the life state, financial memory, and standing actions.

% Trajectory generation uses a directed state-transition graph with admissible precedence, successor, delay, recurrence, precondition, forbidden-state, repeat, and cooldown rules. Conditioning on the persona and current state prevents sequences such as divorce before marriage while permitting branching and recurrence. Registered operations update the persona, state, and standing actions; each trajectory contains 20 event instances.

\begin{table*}[t]
\centering
\scriptsize
\setlength{\tabcolsep}{2.6pt}
\renewcommand{\arraystretch}{1.0}
\caption{Complete life-event ontology and financial-state schema.
Model-visible labels are Korean.}
\label{tab:benchmark-ontologies}
\begin{tabular}{@{}
  >{\raggedright\arraybackslash}p{0.115\textwidth}
  r
  >{\raggedright\arraybackslash}p{0.325\textwidth}
  >{\raggedright\arraybackslash}p{0.115\textwidth}
  r
  >{\raggedright\arraybackslash}p{0.325\textwidth}@{}}
\toprule
\multicolumn{3}{c}{Task~1: life-event ontology (24 types)} &
\multicolumn{3}{c}{Task~2: financial-state schema (34 paths)} \\
\cmidrule(lr){1-3}\cmidrule(l){4-6}
Domain & $n$ & Event identifiers &
Domain & $n$ & State paths \\
\midrule
 
Relationship and household & 7 &
\texttt{marriage, divorce\_or\_separation, childbirth, adoption, dependent\_addition, dependent\_end, family\_death} &
Household & 5 &
\texttt{marital\_status, spouse\_or\_partner, children, dependents, child\_support\_arrangement} \\
\addlinespace[1.5pt]

Housing & 3 &
\texttt{move, home\_purchase, home\_sale} &
Profile & 3 &
\texttt{age, locale, region} \\
\addlinespace[1.5pt]
 
Employment & 6 &
\texttt{employment, reinstatement, job\_change, employment\_end, self\_employment, leave\_of\_absence} &
Employment & 6 &
\texttt{employment\_status, employer, occupation, income\_stability, salary\_day, salary\_account} \\
\addlinespace[1.5pt]
 
Education & 3 &
\texttt{self\_program\_start, child\_stage\_entry, study\_abroad} &
Housing & 9 &
\texttt{residence\_status, address, contract\_type, rent\_amount, rent\_payee, maintenance\_fee\_payee, mortgage\_status, properties, primary\_residence\_property\_id} \\
\addlinespace[1.5pt]
 
Retirement & 2 &
\texttt{start, pension\_start} &
Education & 2 &
\texttt{self\_education\_status, child\_education\_stage} \\
\addlinespace[1.5pt]
 
Crisis & 3 &
\texttt{health\_event, accident\_or\_disaster, financial\_fraud} &
Financial products & 4 &
\texttt{checking\_accounts, savings\_accounts, loans, pension\_or\_irp} \\
\addlinespace[1.5pt]
 
& & &
Financial goals & 4 &
\texttt{emergency\_fund, housing\_deposit\_goal, child\_education\_goal, retirement\_goal} \\
\addlinespace[1.5pt]
 
& & &
Cash flow & 1 &
\texttt{recent\_one\_off\_expense} \\
 
\bottomrule
\end{tabular}
\end{table*}

\subsection{Dialogue Planning and Implementation}

A deterministic planner converts each trajectory into 20 chronological windows of 15 sessions, for 300 sessions in total. Each window corresponds to one occurred event instance and contains exactly one \emph{anchor session}: the first-establishing session for that instance. The remaining positions contain state-compatible non-anchor sessions. This controlled construction standardizes the amount and spacing of observable positive evidence, while the resulting regularity may also serve as a structural cue for reconstruction.

Non-anchor sessions comprise \emph{routine} sessions with no event or state update; \emph{hard negatives} that resemble evidence but preserve the state; \emph{consequence follow-ups} surfacing a downstream result of an established event; \emph{stale-recall follow-ups} contrasting an old value with the current one; and \emph{cancellation evidence} documenting withdrawal of a planned change that is not included in the gold life-event history. These categories test whether models distinguish new occurrences from ordinary activity, near misses, later references, and cancelled plans.

\begin{figure*}[t]
\centering
\includegraphics[width=0.80\textwidth]{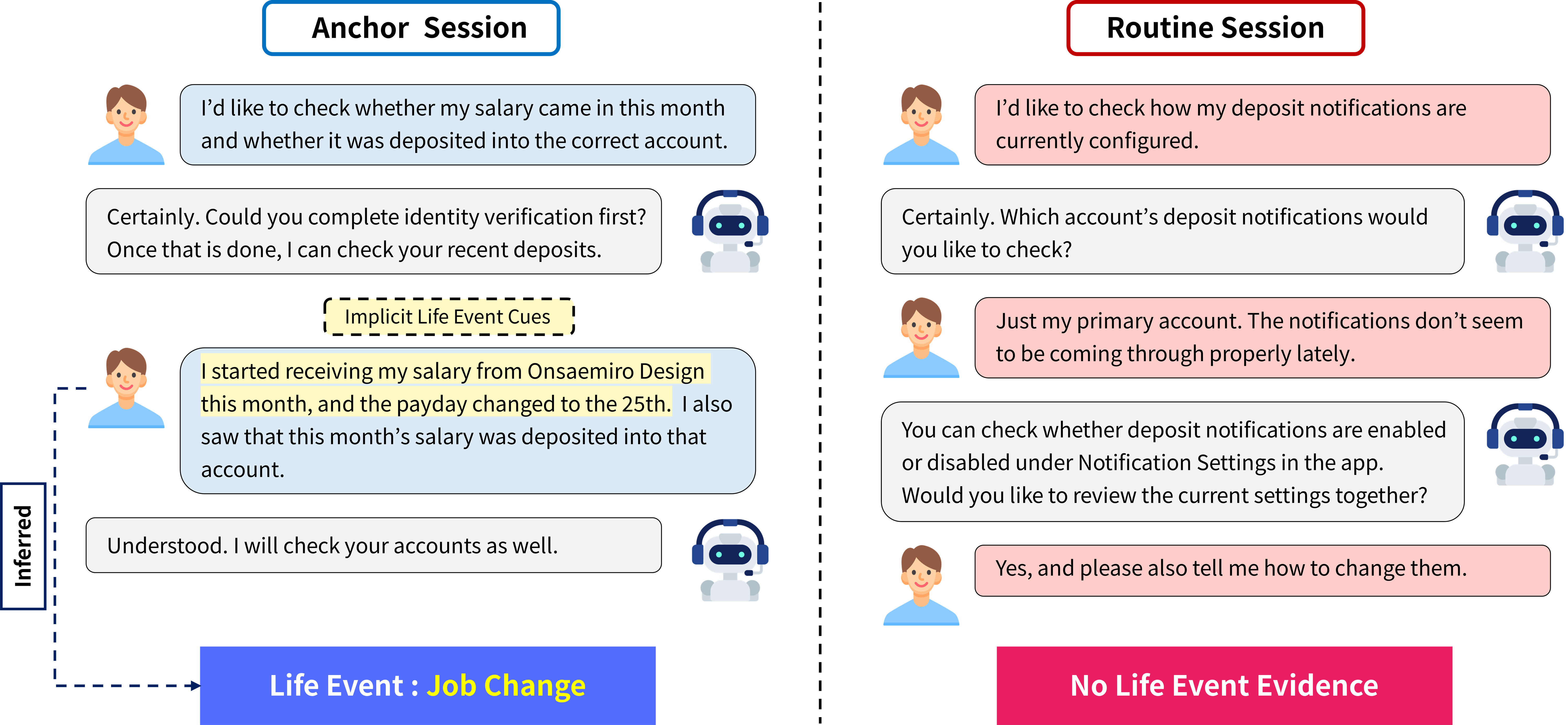}
\caption{An anchor session (left) and a routine session (right). The left dialogue supports a job-change inference from changes in salary source and payday. Visual annotations are excluded from model input. Dialogue is shown in English translation; model input is Korean.}
\Description{Two banking dialogues are shown side by side. The left dialogue highlights a new salary source and payday, with an arrow indicating the inferred job change. The right dialogue concerns deposit notifications and contains no life-event evidence.}
\label{fig:anchor-session}
\end{figure*}

For each session, the planner fixes the banking task, cue placement and grounding, expected state operations, and evaluator-only provenance and safety constraints. \texttt{Claude Sonnet 5} generated each session as an eight-turn mobile- or internet-banking conversation in which event evidence appears incidentally during the customer's banking task. Candidates must pass deterministic schema, grounding, safety, output-contract, and semantic validation; invalid candidates are revised or regenerated within a fixed retry limit. \Cref{fig:anchor-session} contrasts a job-change anchor based on salary-source evidence with a routine session that supports neither a new event nor a state update.

\subsection{Quality Control and Corpus Freezing}
\label{sec:quality-control}

Quality control combines deterministic validation with exhaustive semantic review. Every plan and accepted session is checked for the eight-turn and digital-channel contracts, customer-turn grounding, event recoverability, state-update consistency, hard-negative no-update behavior, assistant leakage, high-risk safety, and surface-form diversity. A complete 300-session trajectory must pass these audits before full production.

All 6,000 sessions were annotated by \texttt{Claude Opus 5} under a fixed seven-criterion rubric. Human effort was concentrated where it carries the most evidential weight rather than spread across exhaustive review. First, an automated screen for near-direct disclosure flagged 180 sessions (3.0\% of the corpus); all 180 were manually revised by three annotators (two Ph.D.-level researchers and one banking service practitioner) before the corpus was frozen. Second, the 400 anchor sessions were each reviewed in full by the same annotators, who assessed evidence leakage, item difficulty, and whether the target state was inferable from the dialogue alone; 15 sessions were revised or regenerated as a result. Domain validity of the rubric and schema was reviewed by three banking-industry practitioners. We release the frozen corpus under CC BY 4.0 together with its annotations, prompts, schema, and scoring code.

% All 6,000 sessions were then evaluated by \texttt{Claude Opus 5} using a fixed seven-criterion rubric. An automated check for near-direct disclosure initially flagged 180 sessions; all were resolved through revision, regeneration, or human adjudication before freezing. Two Ph.D.-level researchers and three banking-industry practitioner then screened the full corpus using the dialogue and evaluator-only metadata, escalating flagged sessions to focused adjudication and resolving disagreements by consensus, for approximately 300 person-hours in total. We will release the frozen corpus under CC BY 4.0 together with its gold annotations, prompts, schema, and scoring code.

\section{Benchmark Tasks}
\label{sec:tasks}

\subsection{Task 1: Grounded Life-Event History Reconstruction}

At each 15-session checkpoint $t \in \{15,30,\ldots,300\}$, Task~1 reconstructs the cumulative grounded life-event history:
\[
\mathcal{H}_t=\{\!\{(e_i,a_i)\}\!\}_{i=1}^{N_t},
\qquad
\widehat{\mathcal{H}}_t=\{\!\{(\widehat e_j,\widehat a_j)\}\!\}_{j=1}^{\widehat N_t}.
\]
Here, $\mathcal{H}_t$ and $\widehat{\mathcal{H}}_t$ denote the gold and predicted multisets of event--anchor pairs, respectively. Each $e_i$ is an occurred life-event type and $a_i$ is the earliest model-visible session whose customer utterances establish that event instance; hats denote predicted quantities, and $N_t$ and $\widehat N_t$ are the gold and predicted numbers of event instances. The model receives all $t$ consecutive sessions through the checkpoint together with the life-event ontology. The prompt neither states $N_t$ nor discloses the one-event-per-window construction rule.

\subsection{Task 2: Complete Financial-State Reconstruction}

At the same checkpoints, Task~2 reconstructs the complete financial state:
\[
\mathcal{S}_t
=
\left\{
p \mapsto (v_{p,t}, z_{p,t}, E_{p,t})
: p \in \mathcal{P}
\right\},
\]
\[
\widehat{\mathcal{S}}_t
=
\left\{
p \mapsto
(\widehat v_{p,t}, \widehat z_{p,t}, \widehat E_{p,t})
: p \in \mathcal{P}
\right\}.
\]
where $\mathcal{P}$ is the set of 34 paths in \Cref{tab:benchmark-ontologies}. For each path $p$, $v_{p,t}$, $z_{p,t}$, and $E_{p,t}$ denote the gold normalized value, validity status, and supporting model-visible session identifiers, respectively; hats denote their predicted counterparts. The request contains the rendered $S_{000}$ state, all model-visible sessions through the checkpoint, all path names, and the machine-readable output schema. Paths whose checkpoint state differs from $S_{000}$ require at least one evidence identifier, whereas paths equal to their $S_{000}$ state require an empty evidence list. Paths with \texttt{unknown} or \texttt{not\_}\allowbreak\texttt{applicable} status may not be omitted, and each checkpoint is evaluated through a fresh request without earlier model predictions.

Gold states use five validity statuses derived from registered operations and revisions: \texttt{current} denotes the latest supported value; \texttt{historical} a superseded value; \texttt{stale} a value that is plausibly invalidated but not replaced; \texttt{unknown} the absence of establishing evidence; and \texttt{not\_}\allowbreak\texttt{applicable} a path excluded by the current state configuration. The model output schema additionally permits \texttt{needs\_verification} as an abstention status.

% The frozen gold derives five statuses from registered operations and revisions: \texttt{current} is latest-supported; \texttt{historical} is superseded; \texttt{stale} is plausibly invalidated but not replaced; \texttt{unknown} lacks establishing evidence; and \texttt{not\_}\allowbreak\texttt{applicable} is excluded by the current configuration. For model-side uncertainty the schema permits \texttt{needs\_verification} as an abstention option; because the frozen gold is fully specified it has zero gold support. It also specifies each path's value type and structure.

\section{Experiments}
\label{sec:experiments}

\subsection{Models and Inference Protocol}

We evaluate eleven LLMs under a \emph{full-context} condition, meaning that the client supplies every dialogue session through the current checkpoint: \texttt{GPT 5.6 Sol}, \texttt{GPT 5.6 Terra}, and \texttt{GPT 5.6 Luna}; \texttt{Claude Opus 4.8} and \texttt{Claude Sonnet 4.6}; \texttt{Gemini 3.1 Pro} and \texttt{Gemini 3.5 Flash}; \texttt{Llama 4 Maverick}; \texttt{GPT-OSS 120B}; \texttt{Qwen 3.5 122B A10B}; and \texttt{Qwen 3.6 35B A3B}. Each model produces 400 outputs per task, for 8,800 predictions in total.

All models receive the same dialogue sessions, taxonomy or schema, and output contract through fresh requests without future sessions or earlier predictions; fallbacks and output repair are disabled. Each of the 8,800 requests contains exactly the sessions visible at its checkpoint. The same input prompt may map to different input lengths across model tokenizer: 125k tokens under Claude tokenizer, 72k under GPT, 66k under Llama, and 64k under Qwen. Peak client-side context utilization is therefore 55.2\% of the published endpoint limit at the tightest endpoint (\texttt{GPT-OSS 120B}, 131,072 tokens) and below 25\% elsewhere; no constructed prefix exceeded the published context limit.

% All eleven models were queried through their native APIs or through a single pinned OpenRouter endpoint between 2026-07-31 and 2026-08-01. No sampling parameters were sent: 

All models were queried between 2026-07-15 and 2026-08-01. We used provider-default sampling and a 20,000-token output cap for every request. Reasoning-capable models used `low' reasoning setting. These settings are provider-defined and not calibrated across model families, so rankings are specific to the evaluated configurations.
 
% The client performed no automatic retries, and the harness allowed at most one parse retry on the OpenRouter profile and none on the direct-API profiles. Reported prefix token counts are each provider's own usage figures and are not comparable across vendors; no single tokenizer was used.

\subsection{Evaluation Metrics}
\label{sec:metrics}

\subsubsection{Task 1 Metrics}

\paragraph{Event--Anchor F1 (EA-F1).}
Event--Anchor F1 (EA-F1, $\uparrow$) is
\begin{equation}
P_t=\frac{|\widehat{\mathcal{H}}_t\cap_{\mathrm{m}}\mathcal{H}_t|}{|\widehat{\mathcal{H}}_t|},\quad
R_t=\frac{|\widehat{\mathcal{H}}_t\cap_{\mathrm{m}}\mathcal{H}_t|}{|\mathcal{H}_t|},\quad
\mathrm{EA\mbox{-}F1}_t=\frac{2P_tR_t}{P_t+R_t},
\end{equation}
where $\cap_{\mathrm{m}}$ denotes multiset intersection. We set $P_t=0$ for an empty prediction; surplus duplicates remain false positives.

\paragraph{Exact History Match (EHM).}
Exact History Match (EHM, $\uparrow$) is
\begin{equation}
\mathrm{EHM}_t=\mathbf{1}\!\left[\widehat{\mathcal{H}}_t=\mathcal{H}_t\right].
\end{equation}

\subsubsection{Task 2 Metrics}

\paragraph{Granular Change Accuracy (GCA@15)}
Because most state paths remain unchanged between adjacent checkpoints, snapshot accuracy alone can obscure whether models apply required updates correctly. We therefore use GCA@15 ($\uparrow$), applying Granular Change Accuracy \citep{GranularChangeAccuracy} to transitions between checkpoints 15 sessions apart. Each path is treated as a slot and its normalized value--status pair as the slot value. GCA distinguishes correct updates from wrong-value, missed, and spurious changes.

\paragraph{Checkpoint State Accuracy (CSA)}
Checkpoint State Accuracy (CSA, $\uparrow$) measures cell-level snapshot correctness:
\begin{equation}
\mathrm{CSA}_t=\frac{1}{|\mathcal{P}|}\sum_{p\in\mathcal{P}}\mathbf{1}\!\left[(\widehat v_{p,t},\widehat z_{p,t})=(v_{p,t},z_{p,t})\right],\qquad |\mathcal{P}|=34.
\end{equation}

\paragraph{Exact Snapshot Match (ESM)}
Exact Snapshot Match (ESM, $\uparrow$) is 1 only when all 34 predicted value--status pairs match gold at a checkpoint.

\paragraph{Evidence Recall (ER)}
GCA defines change relative to the preceding checkpoint, whereas evidence eligibility is defined relative to $S_{000}$. Evidence Recall (ER, $\uparrow$) is computed over the 860 of 1,028 gold transition changes per model whose resulting snapshot carries a gold evidence identifier. The remaining 168 are reversions to their $S_{000}$ value, for which the gold evidence list is empty under the snapshot contract, and are therefore excluded from ER. ER is recall-only and does not penalize surplus citations.

\subsubsection{Shared Metric}

\paragraph{Schema Validity (SV)}
Schema Validity (SV, $\uparrow$) measures compliance with the task-specific output contract. For Task~1, valid outputs must be parseable and conform to the event--anchor schema. For Task~2, valid outputs must be parseable and contain exactly one record for each of the 34 state paths with a valid value type and status.

\subsubsection{Analysis Metrics}

\paragraph{Completeness.}
We analyze checkpoint-wise precision and recall and fit per-model OLS regressions of predicted history size $\widehat N_t$ on gold history size $N_t$.

\paragraph{Type recovery and anchor localization.}
Type-only recall ignores anchors before multiset matching, while conditional anchor accuracy measures exact-anchor accuracy among type-recovered gold instances.

\paragraph{Evidence and temporal-validity diagnostics.}
Predicted session identifiers are joined post hoc to corpus event links to classify cited evidence. For Task~2, we report value, status, and joint accuracy separately.

\paragraph{Cross-task analysis.}
For each model and event instance, evaluation begins at the first checkpoint containing its gold anchor and continues while at least one gold state path remains attributed to that instance; a later overwrite ends the attribution. Task~1 requires the exact event--anchor pair, whereas Task~2 requires the correct value and status on all currently attributed paths, excluding evidence. Of 88,000 candidates, 46,200 are anchor-eligible and 22,583 retain at least one attributed path; zero-path groups are omitted and invalid outputs remain incorrect.

\paragraph{Aggregation and uncertainty.}
Scores are aggregated within trajectories and then across 20 trajectories; GCA@15 uses pooled transition counts. Confidence intervals use 10,000 shared trajectory-cluster percentile-bootstrap resamples (seed 20260725), with checkpoints nested within trajectories. Rank intervals recompute model-level EA-F1, GCA@15, and correlations for the fixed eleven-model set in each resample.

\section{Results}
\label{sec:results}

\begin{table*}[!t]
  % Task 1 narrowed SV is recomputed from final parse errors plus outputs with
  % invalid_prediction_records > 0; ordering-only schema failures are excluded.
  \centering
  \scriptsize
  \setlength{\tabcolsep}{3.2pt}
  \renewcommand{\arraystretch}{1.05}
  \caption{End-to-end results over 400 checkpoints per model and task. Bold and underlined values denote the best and second-best point estimates, respectively. \texttt{Initial Copy} has no life-event history or citations by construction, hence the dashes.}
  \label{tab:main-results}
  \resizebox{0.98\textwidth}{!}{%
    \begin{tabular}{@{}lccccccc@{}}
      \toprule
      & \multicolumn{3}{c}{Task~1: life-event history} & \multicolumn{4}{c}{Task~2: financial state} \\
      \cmidrule(lr){2-4}\cmidrule(l){5-8}
      Model & \makecell{EA-F1 $\uparrow$\\{\scriptsize[95\% CI]}} & EHM $\uparrow$ & SV $\uparrow$ & \makecell{GCA@15 $\uparrow$\\{\scriptsize[95\% CI]}} & CSA $\uparrow$ & ER $\uparrow$ & SV $\uparrow$ \\
      \midrule
      \texttt{Initial Copy} & -- & -- & -- & 0.177 & 0.669 & -- & -- \\
      \midrule
      \texttt{GPT 5.6 Sol} & 0.679 [0.614, 0.740] & 0.068 & 1.000 & 0.413 [0.392, 0.433] & 0.749 & \textbf{0.881} & 0.995 \\
      \texttt{GPT 5.6 Terra} & 0.663 [0.605, 0.720] & 0.065 & 0.993 & 0.411 [0.395, 0.428] & 0.747 & 0.855 & 0.990 \\
      \texttt{GPT 5.6 Luna} & 0.629 [0.576, 0.679] & 0.065 & 0.993 & \underline{0.461} [0.436, 0.484] & \underline{0.771} & 0.786 & 0.978 \\
      \texttt{Claude Opus 4.8} & 0.534 [0.475, 0.586] & 0.050 & 0.983 & \textbf{0.470} [0.443, 0.497] & \textbf{0.801} & 0.848 & 0.965 \\
      \texttt{Claude Sonnet 4.6} & \underline{0.720} [0.658, 0.774] & 0.083 & 0.998 & 0.379 [0.357, 0.402] & 0.711 & \underline{0.866} & 0.970 \\
      \texttt{Gemini 3.1 Pro} & \textbf{0.748} [0.690, 0.799] & 0.113 & 1.000 & 0.438 [0.418, 0.459] & 0.767 & 0.851 & 0.988 \\
      \texttt{Gemini 3.5 Flash} & 0.488 [0.441, 0.533] & 0.053 & 0.995 & 0.412 [0.389, 0.435] & 0.754 & 0.773 & 0.978 \\
      \texttt{Qwen 3.5 122B A10B} & 0.641 [0.580, 0.697] & 0.093 & 1.000 & 0.452 [0.436, 0.469] & 0.681 & 0.664 & 0.990 \\
      \texttt{Qwen 3.6 35B A3B} & 0.473 [0.425, 0.518] & 0.065 & 1.000 & 0.435 [0.422, 0.449] & 0.699 & 0.574 & \underline{0.998} \\
      \texttt{Llama 4 Maverick} & 0.357 [0.309, 0.402] & 0.050 & 0.978 & 0.321 [0.298, 0.344] & 0.672 & 0.393 & 0.963 \\
      \texttt{GPT-OSS 120B} & 0.124 [0.084, 0.167] & 0.018 & 1.000 & 0.249 [0.223, 0.273] & 0.674 & 0.036 & \textbf{1.000} \\
      \bottomrule
    \end{tabular}%
  }
\end{table*}

\subsection{Task 1: Life-Event History Coverage Declines with Depth}
% Source-priority note: review_reanalysis_artifact.json reports cp300 precision
% 0.762205, while direct recomputation from stage1_checkpoint_error_decomposition.csv
% gives 0.764202. We retain the artifact value.
\texttt{Gemini 3.1 Pro} has the highest EA-F1 point estimate (0.748; \Cref{tab:main-results}). Between checkpoints 15 and 300, model-macro precision rises from 0.573 to 0.762, while recall falls from 0.591 to 0.445 and mean per-output EA-F1 falls from 0.579 to 0.532.\footnote{EA-F1 is the mean of per-checkpoint F1, not the F1 of macro precision and recall.} Empty predictions decrease from 62 of 220 outputs (28.2\%) to 5 of 220 (2.3\%), while precision among non-empty outputs remains nearly unchanged (0.80 to 0.78). Thus, models increasingly return some event history but cover a decreasing share of the gold history: underprediction grows from 28.2\% of model--trajectory outputs to 98.2\%. Per-model OLS slopes average 0.524 but vary substantially: high-scoring models have slopes of 0.71--0.85 ($R^2\geq0.82$), whereas the weakest have slopes of 0.05--0.23 ($R^2\leq0.14$).

% Separating identity from grounding isolates the dominant failure. Projecting both multisets onto event identifiers with multiset semantics, pooled type-only recall is 0.533 against 0.462 for full event--anchor pairs. Equivalently, 3,306 of 24,626 type-recovered instances carry the wrong anchor, so conditional anchor accuracy is 0.866.
% % \footnote{Type-only recall uses multiset semantics, so repeated occurrences of one type must be matched in the correct multiplicity; 12,331 of 46,200 Gold instances (26.7\%) are repeats, and crediting a Gold instance whenever its type appears anywhere in the prediction would overstate type-only recall by 11.2 points.} 
% Provenance is therefore not the primary bottleneck: 21,574 of 46,200 gold instances (46.7\%) are missing outright, 6.5 times the 3,306 (7.2\%) attributable to mis-grounding. Across depth, conditional anchor accuracy rises from 0.859 at checkpoint 30 to 0.884 at checkpoint 300 while pair recall falls from 0.580 to 0.445 (\cref{fig:reanalysis-diagnostics}a).

\begin{figure}[t]
  \centering
  \includegraphics[width=\columnwidth]{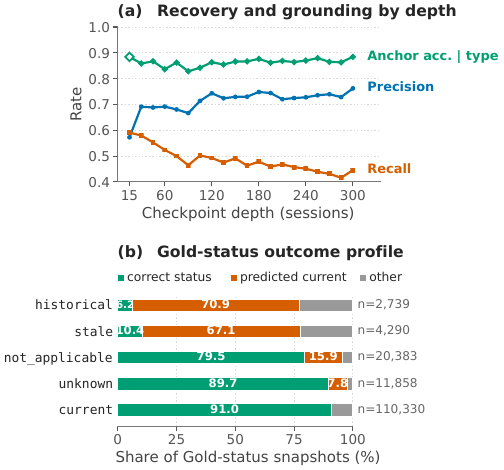}
    \caption{Diagnostics. (a) Model-macro event--anchor precision, recall, and conditional anchor accuracy by checkpoint depth. 
    %At checkpoint 15, gold cardinality is 1 and conditional accuracy is based on 220 observations. 
    Precision is zero for empty predictions. (b) Status outcomes over 149,600 snapshots; gold \texttt{needs\_verification} has no support and is omitted, although models predict it 1,449 times.}
  \Description{Panel (a) shows precision increasing, recall decreasing, and conditional anchor accuracy remaining high across checkpoint depth. Panel (b) shows low historical and stale recall, with both statuses frequently predicted as current.}
  \label{fig:reanalysis-diagnostics}
\end{figure}

Separating event reconstruction from anchor localization shows that coverage is the dominant failure. Pooled type-only recall is 0.533 against 0.462 for full event--anchor pairs. Of 46,200 gold instances, 21,574 (46.7\%) are missing outright, whereas 3,306 (7.2\%) are type-recovered but assigned the wrong anchor, a 6.5-fold difference. Among 24,626 type-recovered instances, conditional anchor accuracy is 0.866. Across depth, it rises from 0.859 at checkpoint 30 to 0.884 at checkpoint 300 even as pair recall falls from 0.580 to 0.445 (\cref{fig:reanalysis-diagnostics}a).

Of the 3,306 wrong-anchor cases, 1,129 (34.2\%) cite a later session linked to the same event occurrence rather than the session that first establishes it. These cases are only 2.4\% of all 46,200 gold instances; the median absolute offset is 14 sessions, about one 15-session window.

Neither distractors nor the strict earliest-session criterion explains much of the gap. Predicted anchors rarely fall on distractor sessions: 90.2\% of predicted pairs point to sessions linked to actual occurred events, hard negatives produce only 33 false positives per 10,000 exposures, and cancellation-anchor sessions are selected just 3 times among 27,467 predicted pairs. Allowing any session linked to the same event occurrence raises anchor accuracy among type-recovered instances only from 0.866 to 0.912.

\subsection{Task 2: State Updates and Temporal Validity}

\texttt{Claude Opus 4.8} attains the highest GCA@15 and CSA point estimates (0.470 and 0.801); ESM peaks at 0.030 because all 34 cells must match. Although 92.44\% of gold transitions are unchanged, \texttt{Initial Copy} reaches 0.669 CSA and 0.010 ESM but only 0.177 GCA@15. Of 11,308 changed transitions, models reconstruct 59.8\% correctly; 18.9\% remain unchanged, 20.5\% are updated incorrectly, and 0.8\% are invalid. Of 138,292 unchanged transitions, 74.0\% are preserved, 13.9\% remain previously wrong, 11.6\% receive spurious updates, and 0.5\% are invalid. Models thus both miss required changes and corrupt stable state.

Across models, pooled errors in financial-state updates are dominated by spurious updates because unchanged cases outnumber changed cases by roughly 12:1, although misses are conditionally more frequent (18.9\% vs.\ 11.6\%).

\Cref{fig:reanalysis-diagnostics}(b) exposes temporal-validity failure masked by pooled accuracy. Across 149,600 snapshots, value, status, and joint accuracy are 79.7\%, 85.4\%, and 73.0\%. Yet \texttt{current} supplies 110,330 cases: always predicting it yields 73.8\% status accuracy, so the observed score adds 11.6 points of lifecycle discrimination. Gold \texttt{historical} and \texttt{stale} recall is only 6.2\% and 10.4\%, with 70.9\% and 67.1\% predicted as \texttt{current}. Joint accuracy exceeds the 68.1\% product-of-marginals reference, indicating concentrated errors.

% Pooled ER ranges from 0.036 to 0.881, led by \texttt{GPT 5.6 Sol}; citing the gold update session does not imply a correct value or status. Failures concentrate by domain: employment is weakest (joint accuracy 0.547, and four of the five weakest individual paths), covering the fields used for income verification, payroll-linked transfers, and credit review.

\section{Analysis}
\label{sec:analysis}

\subsection{Omission and Status Collapse Dominate Mis-grounding}

The dominant failure mode differs by task. On Task~1, event coverage is the main bottleneck: pooled pair recall is 0.462, whereas conditional anchor accuracy reaches 0.866 once the event type is recovered. Every model leaves more gold pairs unmatched than it produces unmatched predictions, with missed:spurious ratios from 1.6:1 to 28.0:1. On Task~2, value recovery is substantially stronger (0.797) than lifecycle tracking, with recall of only 0.062 for \texttt{historical} and 0.104 for \texttt{stale}; pooled state-transition errors are dominated by spurious updates rather than missed changes.

% The diagnostics do not align on one content/structure split. Event identity (0.533) and state values (0.797) are recovered at intermediate rates; pair coverage (0.462) and lifecycle validity (0.062/0.104 for \texttt{historical}/\texttt{stale}) fail sharply, but provenance conditional on identity does not (0.866). On Task~1, models under-record events while usually localizing the designated anchor once an event type is recovered: every one of the eleven leaves more gold pairs unmatched than it emits unmatched predictions (missed:spurious 1.6:1 to 28.0:1). By contrast, pooled state-transition error counts are dominated by spurious updates rather than missed changes.

Conditional anchor accuracy remains high as pair recall falls, unlike the broad evidence-locatability degradation reported in long-context position probes \citep{LostintheMiddle,RULER}. This contrast further indicates that Task~1 failures arise primarily from omitted events rather than inability to localize evidence for recovered events.

% Conditional anchor accuracy remains high while recall falls, unlike uniform evidence-locatability loss in long-context position probes \citep{LostInTheMiddle,RULER}. Its pooled 86.6\% indicates that models usually localize the designated anchor once the event type is recovered.

\subsection{Cross-Task Association Is Sensitive to the Model Set}

Across 22,583 matched observations (2,053 per model), both tasks are correct in 23.6\%, both wrong in 33.2\%, Task~1 only in 24.1\%, and Task~2 only in 19.1\%. Exact grounding neither guarantees correct attributed paths nor is necessary for recovery from later evidence.

For the fixed eleven-model set, EA-F1 and GCA@15 associate weakly: Spearman $\rho = 0.291$ [0.164, 0.409] and Kendall $\tau_b = 0.164$ [0.091, 0.309], computed on pooled GCA@15. The association is carried by two influential points: leave-one-out over all eleven models drops $\rho$ to 0.055 when either of the two weakest models is removed and reaches at most 0.430 otherwise. Mean absolute rank displacement between tasks is 3.1 of 11 positions.

% Opus exceeds Sonnet on GCA@15 by 0.090 [0.067, 0.113] but trails on EA-F1 by 0.186 [0.153, 0.221] (\cref{tab:family-deltas}). 

%Paired contrasts confirm objective-dependent ordering (\cref{tab:family-deltas}).

% \begin{table}[t]
% \centering
% \scriptsize
% \setlength{\tabcolsep}{2.4pt}
% \caption{Focused within-family paired differences (left minus right), with 95\% trajectory-cluster bootstrap intervals.}
% \label{tab:family-deltas}
% \begin{tabular}{@{}lcc@{}}
% \toprule
% Contrast & $\Delta$ EA-F1 & $\Delta$ GCA@15 \\
% \midrule
% Opus $-$ Sonnet & $-$.186 [$-$.221,$-$.153] & .090 [.067,.113] \\
% Sol $-$ Luna & .049 [.016,.081] & $-$.048 [$-$.068,$-$.020] \\
% Pro $-$ Flash & .261 [.218,.304] & .026 [.002,.046] \\
% Qwen 3.5 $-$ 3.6 & .168 [.127,.208] & .017 [.008,.031] \\
% \bottomrule
% \end{tabular}
% \end{table}

% For the fixed eleven-model set, EA-F1 and GCA@15 associate weakly: Spearman $\rho=0.291$ [0.164, 0.409] and Kendall $\tau_b=0.164$ [0.091, 0.309]. The association is carried by two influential points: leave-one-out over all eleven models drops $\rho$ to 0.055 when either of the two weakest models is removed and reaches at most 0.430 otherwise. Mean absolute rank displacement between tasks is 3.1 of 11 positions.

These differences are not explained by output-format failures. SV ranges from 0.978 to 1.000 for Task~1 and from 0.963 to 1.000 for Task~2, yet GCA@15 never exceeds 0.470; perfectly schema-valid \texttt{GPT-OSS 120B} reaches only 0.124 EA-F1, 0.249 GCA@15, and 0.036 ER. Together, the results indicate that event-history and financial-state reconstruction capture distinct aspects of longitudinal reliability.

\section{Limitations}
\label{sec:limitations}

The synthetic benchmark may not reproduce real distributions, ambiguity, or stakes; its 20 independent trajectories and controlled event rates target coverage, not representativeness. Sessions were generated by \texttt{Claude Sonnet 5} and audited by \texttt{Claude Opus 5}, neither of which is among the eleven evaluated models, so no evaluated model was scored on its own generations or judged by itself.
 
We supply static prefixes rather than letting systems manage their own memory, so we measure long-context reconstruction rather than memory-management policy. One event per 15-session window controls density but may reveal gold event cardinality to benchmark-aware systems. Because the dialogue is Korean and language proficiency is not separately controlled, cross-model differences may partly reflect Korean language proficiency. Evidence is scored separately from value and status. Comparisons are observational, and no downstream decision is evaluated.

% Two profile paths are held constant by the generator and are corrected in the released schema. 

\section{Conclusion}
Prior long-horizon benchmarks leave exhaustive, checkpoint-wise
reconstruction of a grounded life-event history and a complete financial
state untested. \textsc{FinLifeBench} separates completeness, provenance,
and temporal validity for this requirement. Across eleven LLMs, the
recurring failures are complementary rather than shared: life-event
histories grow incomplete as the dialogue deepens even though recovered
events are usually anchored to the correct session, whereas
financial-state reconstruction both misses required updates and
overwrites stable cells, and reports superseded or plausibly invalidated
values as current. The best state reconstruction exceeds a
dialogue-blind copy baseline.
Model rankings also differ across the two tasks, reinforcing the value of evaluating both reconstruction objectives.
Even accurately grounded outputs therefore require completeness and
validity checks before they can serve as customer records.

% \section{Conclusion}

% Prior long-horizon benchmarks leave exhaustive, checkpoint-wise reconstruction of a grounded life-event history and a complete financial state untested. \finlifebench\ provides separate measures of completeness, provenance, and temporal validity for this requirement. Across eleven LLMs, recurring failures include incomplete life-event histories and frequent misclassification of \texttt{historical} and \texttt{stale} statuses as \texttt{current}; conditional anchor accuracy reaches 86.6\% once an event type is recovered, while performance on the two objectives is only weakly associated. Thus, even accurately grounded outputs require completeness checks before they can be used as customer records.

\section*{Ethics and Privacy Statement}
The benchmark uses fictional personas and synthetic conversations; no real customer records are involved. The capability it measures---inferring life changes from incidental cues---is itself privacy-sensitive, and systems built on it should obtain consent before updating user profiles and require confirmation before consequential actions. Event rates in the corpus are design parameters, not population statistics, and must not be used for profiling, eligibility, or pricing.

\FloatBarrier

\begin{acks}
This work was supported by
\grantsponsor{iitp}{Institute of Information \& Communications Technology Planning \& Evaluation (IITP)}{https://www.iitp.kr/}
grant funded by the Korea government (MSIT)
(\grantnum{iitp}{RS-2024-00460011},
Climate and Environmental Data Platform for Enhancing Climate Technology Capabilities in the Anthropocene (CEDP))
and (\grantnum{iitp}{RS2021-II211343},
Artificial Intelligence Graduate School Program (Seoul National University)).
This work was also supported by the
\grantsponsor{nrf}{National Research Foundation of Korea (NRF)}{https://www.nrf.re.kr/}
grant funded by the Korea government (MSIT)
(\grantnum{nrf}{RS-2024-00407803}, \grantnum{nrf}{RS-2025-23523657}).
This work was also supported by the BK21 FOUR Program
(Education and Research Center for Industrial Innovation Analytics)
funded by the \grantsponsor{moe}{Ministry of Education, Korea}{https://www.moe.go.kr/}
(No.~\grantnum{moe}{4120240214912}).
This work was also supported by
\grantsponsor{kakaobank}{KakaoBank}{https://www.kakaobank.com/}
under an industry--academia collaboration project
(No.~\grantnum{kakaobank}{0661-20250068}).
\end{acks}

\bibliographystyle{ACM-Reference-Format}
\bibliography{reference}

\end{document}